%% file: neurips_2025.tex
\documentclass{article}

\usepackage[nonatbib,preprint]{neurips_2025}

\usepackage[utf8]{inputenc} 
\usepackage[T1]{fontenc}    
\usepackage{hyperref}       
\usepackage{url}            
\usepackage{booktabs}       
\usepackage{amsfonts}       
\usepackage{nicefrac}       
\usepackage{microtype}      
\usepackage{xcolor}         

\usepackage{graphicx}
\usepackage{amsmath}
\usepackage[numbers]{natbib}
\usepackage{algorithm}
\usepackage{algpseudocode}

\usepackage{tcolorbox}
\usepackage{listings}
\usepackage{rotating} 
\usepackage{courier}
\usepackage{caption}
\usepackage{float}

\title{A 211K Parameter Foundation Model: Genetic Prompting in Neural Cellular Automata}

\author{%
  John Kalkhof \thanks{\texttt{Corresponding author: john.kalkhof@inria.fr}} \quad Niklas Ihm \quad Daniel Kirste \quad Anirban Mukhopadhyay \\[1ex]
  Darmstadt University of Technology, Karolinenplatz 5, 64289 Darmstadt, Germany \\
}

\newcommand{\mymethod}{Foundational-NCA}

\begin{document}

\maketitle

\begin{abstract}

The scale of modern foundation models enables broad task generalization. In medical imaging, where segmentation demands vary widely across clinical contexts, such flexibility is invaluable. But current models, such as SAM trade adaptability for scale, locking their utility behind high-end infrastructure and leaving behind the low-resource environments most in need of diagnostic support. To bridge this gap, we introduce \emph{\mymethod{}}, a lightweight foundation model capable of adapting to a wide range of medical segmentation tasks without retraining. Built on Neural Cellular Automata (NCA), \mymethod{} incorporates \emph{gradient-based learning with genetic optimization over a latent task space}, using a compact task vector to guide predictions. This enables dynamic adaptation via a \emph{genetic prompt mechanism}, allowing the model to generalize across tasks while keeping the parameter count magnitudes lower at \emph{just 211 thousand}, compared to SAM-based models with 38.9 to 636 million parameters. We evaluate \mymethod{} on 15 trained and 6 entirely unseen segmentation tasks from the TotalSegmentator dataset, achieving an average Dice score of 84.5\% on known tasks and 60.4\% on unseen ones. These results demonstrate that task generalization is possible in ultra-compact models, laying the groundwork for resource-efficient, deployable foundation models in clinical environments where infrastructure is limited. By reducing model size by over 180 times, \mymethod{} takes a step toward a more equitable future for medical AI, one where powerful tools are within reach of every healthcare setting.

\end{abstract}

\begin{figure}[tp]
    \centering
    \includegraphics[width=0.85\textwidth]{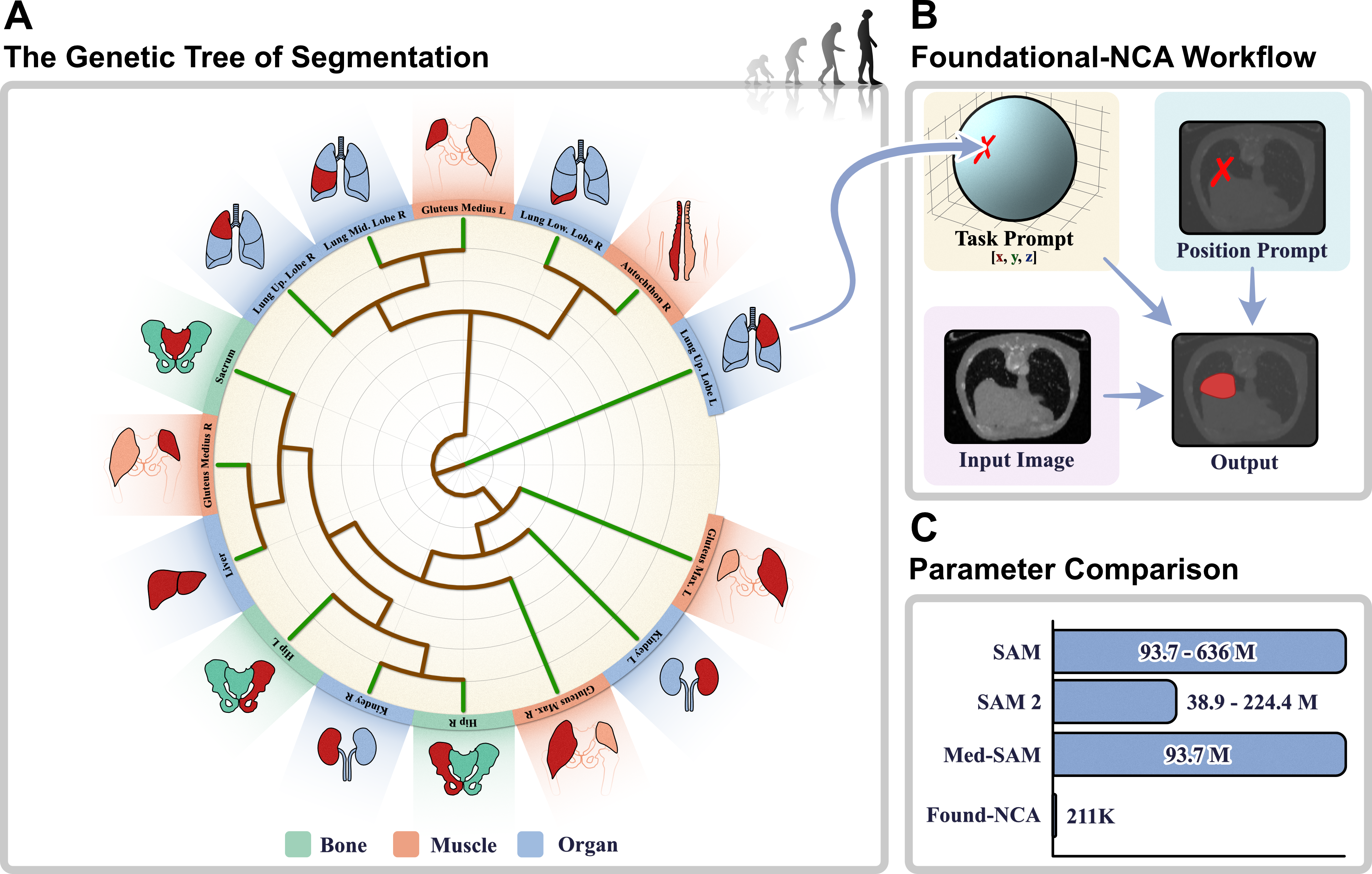}
    \caption{(A) A genetic tree visualizes how \mymethod{} organizes and separates task prompts for different segmentation targets. (B) These task prompts, combined with positional encoding and the input image, guide the model's dynamic prediction process. (C) Unlike SAM variants, which require millions of parameters, \mymethod{} achieves strong generalization with just 211K parameters.}
    \label{fig:Figure1}
\end{figure}

\section{Introduction}

The rise of foundation models has marked a shift from task-specific tools to scalable, unified systems capable of broad generalization. In medical imaging, this versatility promises transformative improvements in diagnostics and patient care \cite{thieme2023foundation}, enabling accurate, scalable segmentation across clinical contexts. Yet, the immense computational scale of these systems, typically involving hundreds of millions of parameters, restricts their deployment to high-resource environments \cite{ajani2021overview}. Consequently, their substantial benefits remain inaccessible precisely in regions where advanced AI could make the greatest difference: low-resource healthcare settings, such as low and middle-income countries with critical shortages of medical personnel \cite{frija2021improve, guo2018application, mccool2022mobile, mollura2020artificial, world2023tracking}.

This disparity has profound implications for global health equity. Accurate and accessible AI-driven segmentation can directly influence patient outcomes, underpinning critical decisions from early disease detection to surgical planning \cite{ma2024segment, rudnicka2024artificial, yao2024cnn}. However, current state-of-the-art solutions, such as the Segment Anything Model (SAM) \cite{kirillov2023segment} and its medically-adapted variant MedSAM \cite{ma2024segment}, remain impractically large, requiring between 38.9 and 636 million parameters. This complexity makes their deployment on widely available, cost-effective hardware such as smartphones infeasible, particularly problematic in medical settings where strict privacy regulations prevent the transfer of sensitive patient data to external servers \cite{loftus2022federated, teo2024federated}.

Neural Cellular Automata (NCAs) \cite{gilpin2019cellular, mordvintsev2020growing}, inspired by self-organizing biological systems, present a promising alternative due to their minimal resource requirements \cite{kalkhof2024unsupervised}, low parameter count \cite{deutges2024neural, kalkhof2023med, kalkhof2025parameter, mordvintsev2021mu, ranem2024nca, ranem2025ncadapt} and inherent adaptability \cite{kalkhof2025med}. Recent advancements, such as Med-NCA \cite{kalkhof2023med, kalkhof2025med}, have shown that NCA can achieve strong segmentation performance with significantly fewer resources. However, current NCA-based segmentation models are fundamentally limited: \emph{they are designed for single-task scenarios} and lack the flexibility required to handle multiple tasks dynamically.

In this work, we introduce \mymethod{}, the first NCA-based foundation model capable of adapting dynamically across multiple segmentation tasks via prompt-based conditioning. Central to our approach is a compact, continuous \emph{task vector}, optimized using \emph{genetic algorithms} (GA) \cite{sampson1976adaptation} to encode specific segmentation objectives. This task vector modulates the NCA's update dynamics at each step, allowing a single fixed model to perform diverse segmentation tasks without any weight modifications (Figure \ref{fig:Figure1}). By integrating genetic optimization for task conditioning with gradient-based training of model parameters, \mymethod{} generalizes effectively across \emph{15 trained anatomical targets} and demonstrates promising adaptability to \emph{6 entirely unseen segmentation tasks}, delivering foundation-level versatility with only \emph{211K parameters}.

Evaluated on the TotalSegmentator dataset \cite{wasserthal2023totalsegmentator}, \mymethod{} achieves an average Dice score of 84.5\% across 15 training tasks and 60.4\% on 6 unseen tasks, using only 211K parameters. In comparison, MedSAM, a 93-million-parameter foundation model, attains 75.8\% and 73.4\% on the same task sets. While \mymethod{} shows a trade-off in generalization to unseen targets, it outperforms MedSAM on seen tasks with over $400\times$ fewer parameters, highlighting the potential of compact, prompt-driven models for low-resource scenarios.

By substantially reducing hardware requirements and demonstrating a first step toward generalization within an ultra-lightweight foundation model, \mymethod{} offers a promising direction for democratizing medical AI. It provides an initial framework toward universally accessible foundation models, compact, adaptive, and with the potential to bring advanced healthcare solutions closer to everyone, everywhere.


\section{Related Work}
This work builds upon two key areas: \emph{genetic optimization}, a class of gradient-free search algorithms well-suited to non-differentiable objectives, and \emph{neural cellular automata}, a paradigm for self-organizing, locally updated neural networks.

\subsection{Genetic Optimization}

Genetic Optimization (GO), originally introduced in the 1970s \cite{sampson1976adaptation}, simulates evolutionary processes by iteratively improving a population of candidate solutions through selection, mutation, and crossover. Given an objective function $f: \mathbb{R}^d \rightarrow \mathbb{R}$, GO seeks an optimal input $x^* \in \mathbb{R}^d$ such that $f(x^*) = \min_{x \in \mathbb{R}^d} f(x)$, without requiring $f$ to be differentiable or continuous.

Let $\mathcal{P}_t = \{x_i^{(t)}\}_{i=1}^N \subset \mathbb{R}^d$ denote a population at generation $t$. Each candidate $x_i^{(t)}$ is evaluated via the scalar fitness function $f(x_i^{(t)})$. A selection operator $\mathcal{S}$ retains the top $k < N$. New candidates are then sampled via a crossover operator $\mathcal{C}$ and perturbed through mutation $\mathcal{M}$, i.e.,

$$
x_j^{(t+1)} = \mathcal{M}(\mathcal{C}(x_{a}^{(t)}, x_{b}^{(t)}), \epsilon), \quad x_{a}, x_{b} \in \mathcal{S}(\mathcal{P}_t),
$$

where $\epsilon \sim \mathcal{N}(0, \sigma^2 I)$ introduces stochastic variation. The process repeats until convergence or a fixed budget is reached.

The strength of GO lies in its independence from gradient information, enabling optimization over black-box, discrete, or non-differentiable objectives. Its main drawback is sample inefficiency: convergence requires orders of magnitude more evaluations than gradient-based methods \cite{mitchell1998introduction, eiben2015introduction}.

\subsection{Neural Cellular Automata}

Neural Cellular Automata (NCA) are models characterized by local communication and self-organization, first introduced by \cite{mordvintsev2020growing}. They build upon the concept of classical Cellular Automata, originally proposed by von Neumann \cite{von1966theory} in the 1940s, where complex global behavior emerges from simple local interactions. Unlike traditional Cellular Automata, which rely on hand-crafted update rules, NCA replace these rules with a \emph{learnable, fully convolutional neural network with pixel-based dropout}, that governs the state updates of each cell based solely on local neighborhood information.

The model evolves iteratively, applying the learned update rule over multiple time steps to reach a predefined target state. However, early applications of NCA were largely confined to toy problems due to the high memory demands of storing state trajectories during training and slow local communication. This bottleneck was alleviated by the work of \cite{kalkhof2023med}, who introduced a memory-efficient training procedure that allows NCA to be applied to more complex and practical tasks.

\begin{figure}[t]
    \centering
    \includegraphics[width=1.0\textwidth]{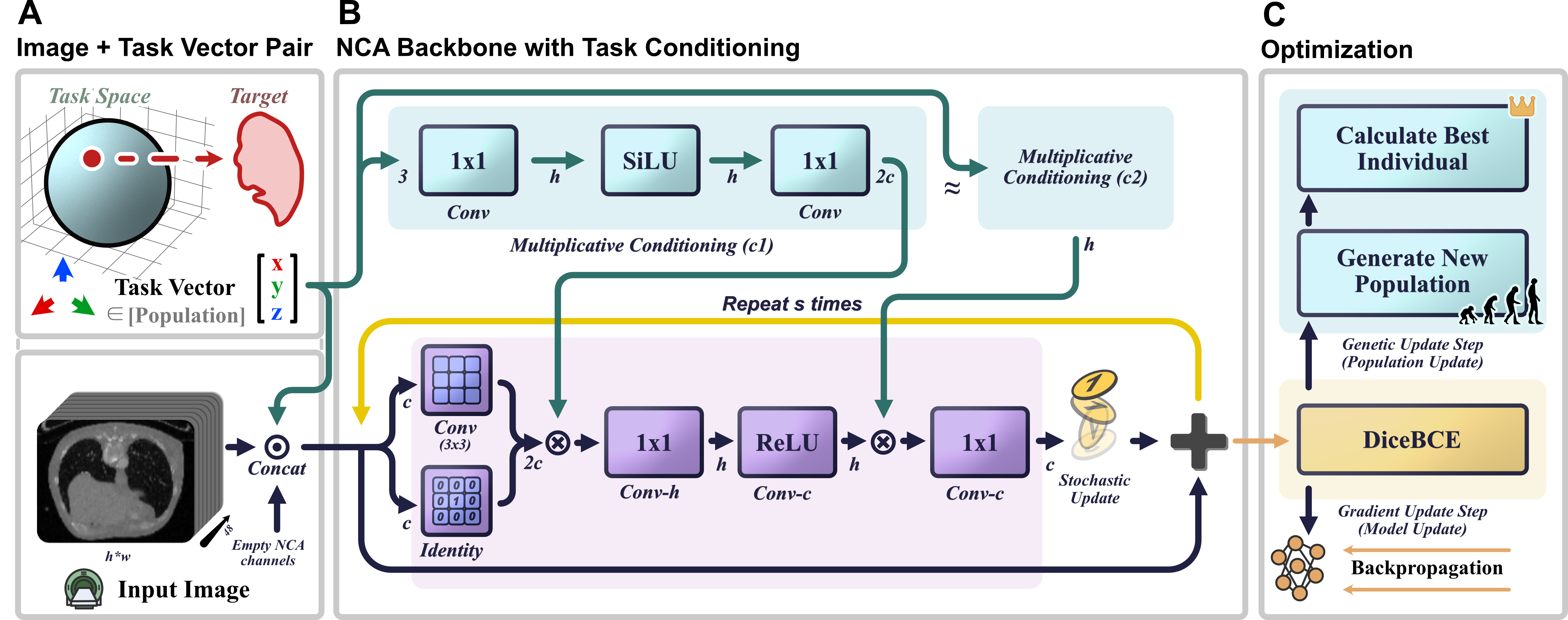}
    \caption{(A) A single input image is paired with a task-specific latent vector, enabling the model to produce different segmentation outputs for the same input. (B) The task vector serves as a multiplicative conditioning signal within the NCA architecture, modulating the update dynamics. (C) Training alternates between two phases: model updates are performed via gradient descent, while task vector optimization is carried out using genetic optimization.}
    \label{fig:genca_figure}
\end{figure}

\section{Methodology}
We introduce \mymethod{}, a compact NCA architecture conditioned on a learned task vector, enabling segmentation of multiple anatomical targets without weight modification. This is achieved through a combination of gradient-based training and genetic optimization.

\subsection{Backbone}
We build on the \emph{Med-NCA} architecture \cite{kalkhof2023med}, which enhances NCA for high-resolution medical image segmentation by addressing two key challenges: (1) the inability of standard NCAs to capture both global context and fine spatial detail, and (2) the high memory cost of storing recurrent NCA states during training. Med-NCA introduces a multi-scale hierarchical design, where a coarse-resolution model $m_1$ processes a downsampled image, and a fine-resolution model $m_2$ refines the output using both high-resolution input and coarse features. This design enables the model to reason over global structures while preserving fine anatomical boundaries. Additionally, a randomized patch-based training strategy greatly reduces GPU memory requirements. We refer readers to the Med-NCA papers \cite{kalkhof2023med, kalkhof2025med} for full architectural and implementation details.

\subsection{Prompting NCAs}
In \mymethod{}, we extend Med-NCA by incorporating two conditioning prompts: a \emph{task prompt} and a \emph{positional prompt}. These prompts are integrated into the NCA architecture to provide additional contextual information during the segmentation process. The task prompt specifies the segmentation target, while the positional prompt defines a starting point, allowing the model to better localize anatomical regions.

To integrate the prompts into the NCA architecture, the task prompt is concatenated directly to the state of each cell in the NCA grid (Figure \ref{fig:genca_figure} A), ensuring task-specific context is available throughout the update process. Additionally, the task prompt is processed by two small networks consisting of $1 \times 1$ convolutions and a \texttt{SiLU} activation, with its output applied multiplicatively to two intermediate feature maps during each NCA update (Figure \ref{fig:genca_figure} B). For simplicity, Figure \ref{fig:genca_figure} shows the architecture for one level, but the same process is applied to both the coarse and fine-resolution models $m_1$ and $m_2$. Assuming a meaningful task representation is stored in the task space, the model can dynamically adapt to different segmentation tasks by simply modifying the task vector prompt, without the need for retraining.

\subsection{Genetic Task Vector}

We propose a novel formulation where each input image $x \in \mathbb{R}^{H \times W \times C}$ is paired with a latent \emph{task vector} $t \in \mathbb{R}^d$, which specifies the segmentation target. The task vector $t$ allows the model to adapt to different segmentation tasks. Unlike traditional multi-class segmentation models, we do not provide explicit task labels or predefined class embeddings.

To optimize $t$ for a given [image $x$, mask $y$] pair, we use \emph{genetic optimization} (GO), which iteratively refines a task vector Pool $\mathcal{P}_t(x)$ based on segmentation performance (Algorithm \ref{alg:task_opt}). This process converges to a region in task space that yields high segmentation performance for the target anatomy, \emph{without requiring the model to be explicitly told what the anatomy is}.

\begin{algorithm}[h!]
\caption{Genetic Task Population Optimization (Single Step)}
\label{alg:task_opt}
\begin{algorithmic}[1]
\State \textbf{Input:} Image $x$, ground truth mask $y$, current population $\mathcal{P}_t(x)$
\State Evaluate fitness using DiceBCE between predicted and ground truth segmentation
\State Select top-k candidates from $\mathcal{P}_t$
\State Generate new population by mutation and interpolation:

$$
t_j^{(g+1)} = \alpha t_a^{(g)} + (1-\alpha) t_b^{(g)} + \epsilon, \quad \epsilon \sim [-1, 1]
$$

\State Project task vectors onto unit sphere: $t_j^{(g+1)} \leftarrow t_j^{(g+1)} / \| t_j^{(g+1)} \|_2$
\State \textbf{Output:} Optimized task vector $\mathcal{P}_t(x)^*$
\end{algorithmic}
\end{algorithm}

This optimization enables the NCA to decouple the concepts of \emph{what to segment} and \emph{how to segment}, facilitating foundation-like generalization within a compact architecture. Once an optimal task vector $t^*$ is identified for a class through genetic optimization, it \emph{can be reused for other images of the same class at inference time}. Importantly, \textbf{we discover task vectors for unseen tasks by optimizing $t$ using only a single annotated scan} from a held-out validation patient, enabling model reuse with minimal supervision and efficient segmentation across unseen tasks.

\subsection{Alternating Genetic and Gradient-Based Optimization}
We use an alternating training scheme, combining gradient descent, optimized with Adam \cite{kingma2014adam}, and genetic optimization using the PyGAD library \cite{gad2024pygad}, to jointly optimize model parameters and task vectors. The process is described in Algorithm \ref{alg:alternating}. 

\begin{algorithm}[h!]
\caption{Alternating Genetic and Gradient-Based Optimization}
\label{alg:alternating}
\begin{algorithmic}[1]
\State \textbf{Input:} $\mathcal{D} = \{(x_i, y_i, \mathcal{P}_i^{(0)})\}_{i=1}^n$, initial model parameters $\theta$
\For{epoch = 1 to N = 200}
    \If{epoch <= 20}
        \State Train model using gradient descent (fix task vector $\mathbf{t}$)
    \ElsIf{epoch = 21}
        \State Initialize task vector $\mathbf{t}$ from task pool (sample and project onto unit sphere)
        \State Run genetic optimization: evaluate, generate offspring, and select best task vector
    \Else
        \State Continue gradient descent with updated task vector $\mathbf{t}$
        \If{epoch \% interval == 0}
            \State Perform genetic optimization to refine $\mathbf{t}$ using Algorithm \ref{alg:task_opt}.
        \EndIf
        \State Linearly increase genetic optimization interval (from 8 to 16 epochs)
    \EndIf
\EndFor
\State \textbf{Output:} Optimized model parameters $\theta$ and task vectors $\mathbf{t}$
\end{algorithmic}
\end{algorithm}

\subsection{Two-stage training}

To improve training stability and prompt utilization, we adopt a two-stage training strategy. We observed that when both the \emph{task prompt} $\mathbf{t}$ and the \emph{positional prompt} $\mathbf{p}$ are introduced from the start, the model tends to ignore $\mathbf{t}$, as it is initialized to zero and only gradually refined. This limits the amount of information that can be encoded in the task vector. Conversely, if $\mathbf{p}$ is added later, it is similarly neglected, since the model has already converged without relying on positional information.

Instead, we first train the model using only $\mathbf{t}$, allowing it to optimize a meaningful task embedding space via standard supervision $\mathcal{L}(f_\theta(\mathbf{x}, \mathbf{t}), \mathbf{y})$ (see Algorithm \ref{alg:alternating}). After this phase, we reset the model weights $\theta$, retain the learned task prompts $\mathbf{t}$, and begin training with both $\mathbf{t}$ and $\mathbf{p}$, this time keeping $\mathbf{t}$ fixed, hence repeating Algorithm \ref{alg:alternating} without the genetic optimization. This ensures the model learns to effectively leverage both prompts from the start, leading to more consistent convergence and better overall performance.

\section{Experimental Setup and Results}
We evaluate \mymethod{} on segmentation accuracy, generalization to unseen tasks, and the structure of its learned task space. Implemented in PyTorch \cite{paszke2019pytorch}, our experiments benchmark performance against state-of-the-art foundation models while highlighting the benefits of prompt-based control in a compact architecture.

\subsection{Datasets}
We evaluate our method on the TotalSegmentator dataset \cite{wasserthal2023totalsegmentator}, which provides 1,228 CT scans with annotations for 117 anatomical structures. For training, we select \emph{15 medium- to large-sized targets}, including the gluteus maximus and medius (left and right), hips, kidneys, liver, sacrum, the right autochthon muscle, and lobes of the right and left lung. To assess generalization, we evaluate on \emph{six anatomies not seen during training}: the left autochthon and iliopsoas muscles, left lower lung lobe, sternum, stomach, and urinary bladder. We exclude low-quality scans where masks extend beyond the field of view or which show severe scaling artifacts, and randomly select 100 patients for testing, 5 for validation, and 300 slices per task of the remaining patients for training (in total 4500 slices). All 3D scans are processed as 2D axial slices resized to $164\times164$, with evaluation performed at the patient level across all slices. Slices lacking positional prompts are treated as empty predictions.

\begin{table}[htbp]
\centering
\caption{Dice accuracy comparison of \mymethod{} against SAM, SAM2 and MedSAM on Task Set A (tasks seen during training) and Task Set B (unseen during training).}
\begin{tabular}{lcccc}
\toprule
\textbf{Task Set A} & \textbf{\mymethod{}} & \textbf{SAM} & \textbf{SAM 2} & \textbf{Med-SAM}\\
\emph{Seen} & \emph{211K param.} & \emph{93.7M param.} & \emph{38.9M param.} & \emph{93.7M param.}\\
\midrule
Gluteus Max. L & 0.885 $\pm$ 0.077 & 0.054 $\pm$ 0.014 & 0.073 $\pm$ 0.019 & 0.814 $\pm$ 0.057\\
Gluteus Max. R & 0.874 $\pm$ 0.073 & 0.076 $\pm$ 0.021 & 0.075 $\pm$ 0.020 & 0.796 $\pm$ 0.064\\
Liver & 0.911 $\pm$ 0.105 & 0.343 $\pm$ 0.076 & 0.369 $\pm$ 0.092 & 0.891 $\pm$ 0.068\\
Lung Lower Lobe R & 0.832 $\pm$ 0.162 & 0.638 $\pm$ 0.163 & 0.552 $\pm$ 0.174 & 0.809 $\pm$ 0.116\\
Lung Middle Lobe R & 0.808 $\pm$ 0.114 & 0.463 $\pm$ 0.125 & 0.312 $\pm$ 0.109 & 0.842 $\pm$ 0.080\\
Lung Upper Lobe L & 0.870 $\pm$ 0.170 & 0.559 $\pm$ 0.228 & 0.457 $\pm$ 0.216 & 0.866 $\pm$ 0.102\\
Lung Upper Lobe R & 0.800 $\pm$ 0.271 & 0.628 $\pm$ 0.225 & 0.501 $\pm$ 0.210 & 0.841 $\pm$ 0.202\\
Autochthon R & 0.811 $\pm$ 0.060 & 0.032 $\pm$ 0.010 & 0.034 $\pm$ 0.010 & 0.709 $\pm$ 0.033\\
Kidney L & 0.840 $\pm$ 0.186 & 0.425 $\pm$ 0.320 & 0.404 $\pm$ 0.281 & 0.798 $\pm$ 0.159\\
Kidney R & 0.871 $\pm$ 0.153 & 0.420 $\pm$ 0.296 & 0.418 $\pm$ 0.273 & 0.826 $\pm$ 0.071\\
Sacrum & 0.758 $\pm$ 0.104 & 0.237 $\pm$ 0.139 & 0.159 $\pm$ 0.117 & 0.600 $\pm$ 0.082\\
Gluteus Medius L & 0.828 $\pm$ 0.124 & 0.054 $\pm$ 0.014 & 0.053 $\pm$ 0.014 & 0.674 $\pm$ 0.102\\
Gluteus Medius R & 0.840 $\pm$ 0.047 & 0.049 $\pm$ 0.013 & 0.046 $\pm$ 0.011 & 0.709 $\pm$ 0.064\\
Hip L & 0.863 $\pm$ 0.101 & 0.440 $\pm$ 0.208 & 0.357 $\pm$ 0.195 & 0.593 $\pm$ 0.102\\
Hip R & 0.887 $\pm$ 0.104 & 0.449 $\pm$ 0.209 & 0.345 $\pm$ 0.209 & 0.596 $\pm$ 0.116\\
\midrule
Average & 0.845 & 0.325 & 0.277 & 0.758\\
\midrule
\midrule
\textbf{Task Set B} & \textbf{\mymethod{}} & \textbf{SAM} & \textbf{SAM 2} & \textbf{Med-SAM}\\
\emph{Unseen} & \emph{211K param.} & \emph{93.7M param.} & \emph{38.9M param.} & \emph{93.7M param.}\\
\midrule
Autochthon L & 0.678 $\pm$ 0.159 & 0.032 $\pm$ 0.010 & 0.034 $\pm$ 0.011 & 0.713 $\pm$ 0.032\\
Iliopsoas L & 0.568 $\pm$ 0.174 & 0.035 $\pm$ 0.050 & 0.025 $\pm$ 0.018 & 0.627 $\pm$ 0.150\\
Lung Lower Lobe L & 0.730 $\pm$ 0.153 & 0.570 $\pm$ 0.204 & 0.546 $\pm$ 0.216 & 0.791 $\pm$ 0.119\\
Sternum & 0.661 $\pm$ 0.088 & 0.191 $\pm$ 0.236 & 0.039 $\pm$ 0.099 & 0.685 $\pm$ 0.100\\
Stomach & 0.492 $\pm$ 0.184 & 0.126 $\pm$ 0.070 & 0.152 $\pm$ 0.108 & 0.785 $\pm$ 0.066\\
Urinary Bladder & 0.497 $\pm$ 0.179 & 0.132 $\pm$ 0.104 & 0.151 $\pm$ 0.123 & 0.803 $\pm$ 0.092\\
\midrule
Average & 0.604 & 0.181 & 0.158 & 0.734\\

\bottomrule
\end{tabular}
\label{tab:gluteus_max_l}
\end{table}

\subsection{Quantitative Results}
Table \ref{tab:gluteus_max_l} reports the Dice scores of \mymethod{} compared to several foundation models, including SAM (Version B), SAM 2 (Version 2.1-T), and MedSAM, across both Task Set A (seen classes) and Task Set B (unseen classes). On Task Set A, \mymethod{} achieves the highest average performance (0.845 Dice), \emph{outperforming all three foundation models}, MedSAM (0.758 Dice) and SAM (0.325 Dice) and SAM 2 \cite{ravi2024sam2} (0.277) by a substantial margin. Despite being two to three orders of magnitude smaller in parameter count (211k vs. 93M), this demonstrates that task-specific supervision, even in compact architectures, can exceed the performance of large, general-purpose foundation models when applied to domain-specific data.

On Task Set B, which includes anatomies not used during training, \mymethod{} generalizes reasonably well (average Dice: 0.604), substantially outperforming SAM (0.181 Dice) and SAM 2 (0.158 Dice), but trailing MedSAM (0.734 Dice), which benefits from training on over 11 million images compared to just 4,500 slices for \mymethod{}. For each unseen class, \mymethod{} is evaluated five times using a different randomly selected validation patient to optimize the task prompt (takes 26.1 Minutes), ensuring a robust estimate of generalization. Given the parameter disparity, these results illustrate the efficiency of our architecture in achieving competitive generalization at a fraction of the parameter requirements. In particular, performance is strongest on structurally similar or overlapping anatomies (e.g., lungs, autochthon), but weaker on highly variable or deformable targets like the stomach or bladder, suggesting the limits of task adaptation when the latent space has not been exposed to sufficient anatomical diversity during training.

Looking ahead, we anticipate that increasing task diversity by leveraging larger datasets with hundreds of anatomical structures will enrich the latent task representation. This would improve generalization, enabling the model to adapt more accurately to unseen tasks and approach full foundation-model performance with a fraction of the parameter budget.

\begin{figure}[htbp]
    \centering
    \includegraphics[width=0.8\textwidth]{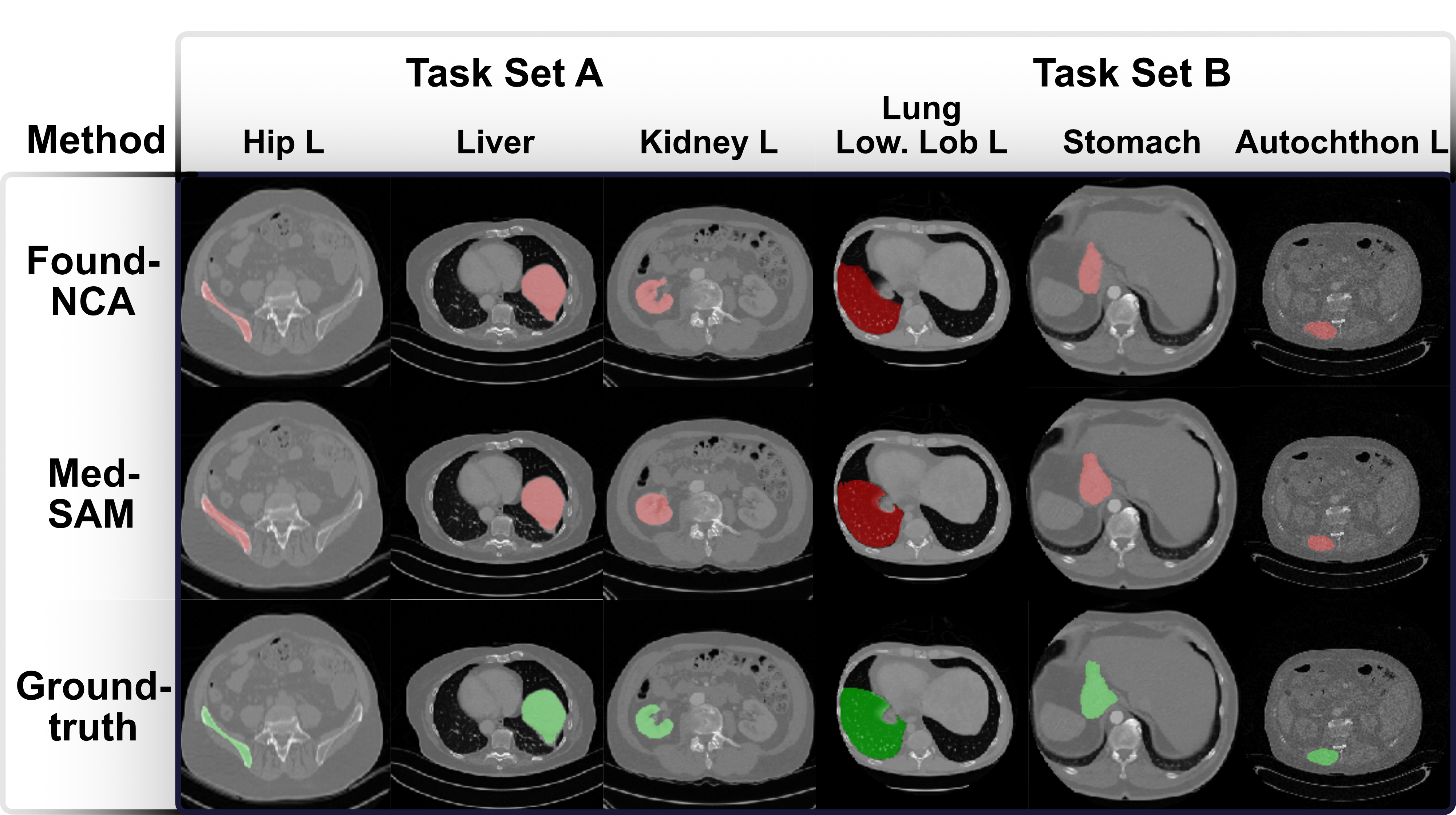}
    \caption{Exemplary segmentation results comparing \mymethod{} with Med-SAM.}
    \label{fig:qualitative}
\end{figure}

\subsection{Qualitative Results}

The qualitative results in Figure \ref{fig:qualitative} demonstrate \mymethod{}'s strong performance on seen segmentation tasks. For tasks such as hippocampus left, \mymethod{} accurately captures the target shape, and for the hip, it follows bone contours precisely. Despite being trained on a smaller dataset of 4,500 images compared to Med-SAM’s 11 million, \mymethod{} still shows strong adaptability to unseen tasks. While it struggles with finer details, it effectively fills the target regions, and the general shape of the anatomical structures is well-preserved. This indicates that \mymethod{} can successfully adapt to new tasks and generalize to previously unseen data. With more data and further training, \mymethod{} has the potential to close the gap in detail and achieve similar performance across a wider range of tasks.

\begin{table}[htbp]
\centering
\caption{Ablation study of \mymethod{} on unseen tasks, evaluating the impact of removing the task vector and the positional prompt. Results for seen tasks are provided in the supplementary.}
\begin{tabular}{lccc}
\toprule
\textbf{Task Set A} & \textbf{\mymethod{}} & \textbf{No Location} & \textbf{No Task}  \\
Average & 0.845 & 0.753 & 0.848 \\
\midrule
\textbf{Task Set B} & \textbf{\mymethod{}} & \textbf{No Location} & \textbf{No Task}  \\
\midrule
Autochthon L & 0.678 $\pm$ 0.159 & 0.002 $\pm$ 0.012 & 0.661 $\pm$ 0.148\\
Iliopsoas L & 0.568 $\pm$ 0.174 & 0.020 $\pm$ 0.041 & 0.593 $\pm$ 0.141\\
Lung Lower Lobe L & 0.730 $\pm$ 0.153 & 0.299 $\pm$ 0.168 & 0.746 $\pm$ 0.158\\
Sternum & 0.661 $\pm$ 0.088 & 0.018 $\pm$ 0.038 & 0.645 $\pm$ 0.079\\
Stomach & 0.492 $\pm$ 0.184 & 0.019 $\pm$ 0.046 & 0.392 $\pm$ 0.152\\
Urinary Bladder & 0.497 $\pm$ 0.179 & 0.036 $\pm$ 0.058 & 0.271 $\pm$ 0.139\\
\midrule
Average & 0.604 & 0.066 & 0.551 \\

\bottomrule
\end{tabular}
\label{tab:ablation}
\end{table}

\subsection{Ablation}
To assess the role of each component in \mymethod{}, we perform an ablation study on the unseen task set (Table~\ref{tab:ablation}), evaluating variants without the \textbf{task vector} ("No Task") and without the \textbf{positional prompt} ("No Location"). Results for seen tasks are in the supplementary material. Removing the positional prompt causes a steep performance drop on unseen tasks (average Dice: 0.070), indicating its essential role in spatial localization. Removing the task vector yields moderate degradation (Dice: 0.551), with especially large losses on anatomically variable structures like the stomach and bladder. Together, these results show that the positional prompt guides \emph{where} to segment, while the task vector helps determine \emph{what} to segment. Both are critical for enabling \mymethod{} to generalize across tasks within such a compact architecture.

\begin{figure}[htbp]
    \centering
    \includegraphics[width=0.95\textwidth]{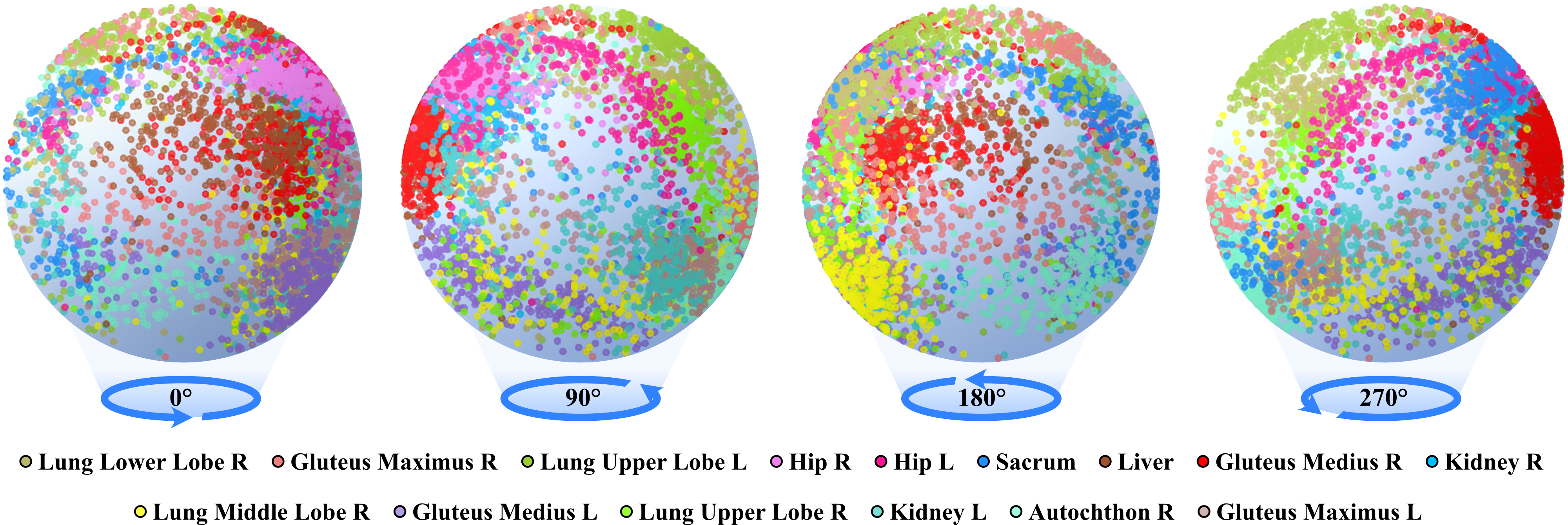}
    \caption{The four globes illustrate how automatically optimized task vectors align with the tasks they are intended to solve. Each globe presents the task sphere from a different viewing angle to highlight the spatial distribution of task representations.}
    \label{fig:taskvector_globe}
\end{figure}

\subsection{Task Space Organisation}
Figure \ref{fig:taskvector_globe} shows how the task space of \mymethod{} becomes organized through training, despite the model never receiving supervision about task identities. Instead, genetic optimization iteratively adjusts each task vector to maximize segmentation accuracy relative to the ground truth. For visualization, each learned task vector is colored according to the target anatomy it ultimately segments, although this label is never used during training. The result reveals distinct clustering in task space: \emph{anatomies with similar shape or location tend to occupy nearby regions}. This emergent structure highlights the model’s ability to \emph{disentangle and organize tasks purely through the interplay of gradient descent and genetic optimization}, even in the absence of explicit class information.

\begin{figure}[htbp]
    \centering
    \includegraphics[width=0.95\textwidth]{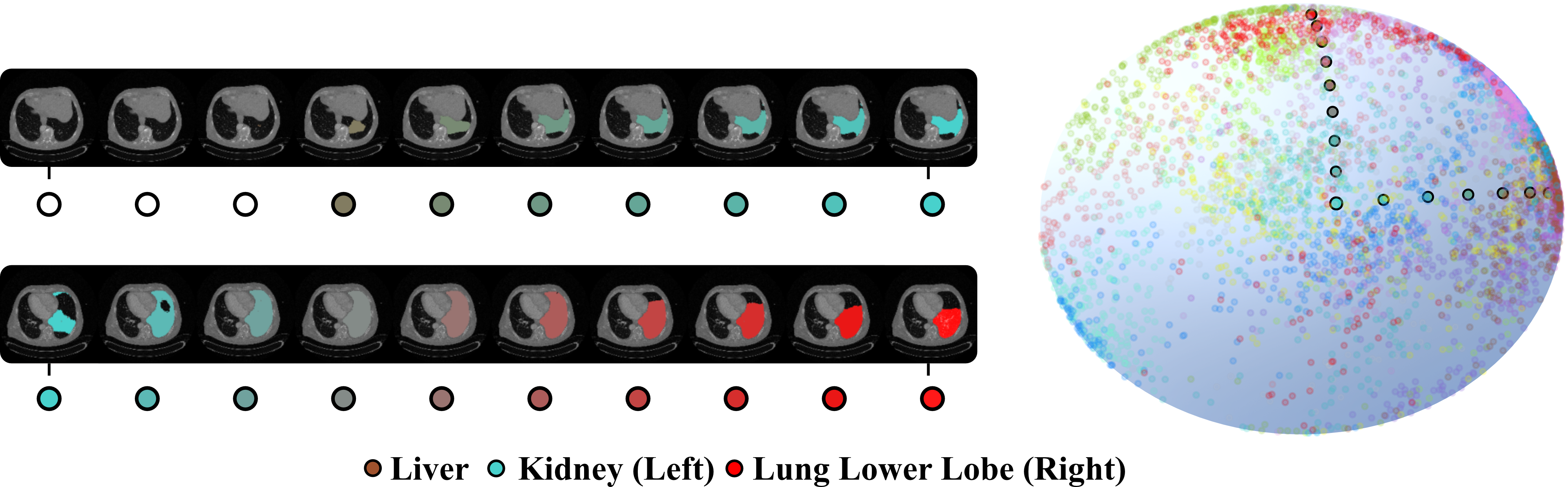}
    \caption{Interpolation between task vectors on the spherical task space. In both liver-to-kidney and kidney-to-lung transitions, the model gradually morphs the predicted segmentation shape to match the evolving task prompt while adapting it to the fixed anatomical position.}
    \label{fig:taskvector_interpolation}
\end{figure}

\begin{figure}[htbp]
    \centering
    \includegraphics[width=0.95\textwidth]{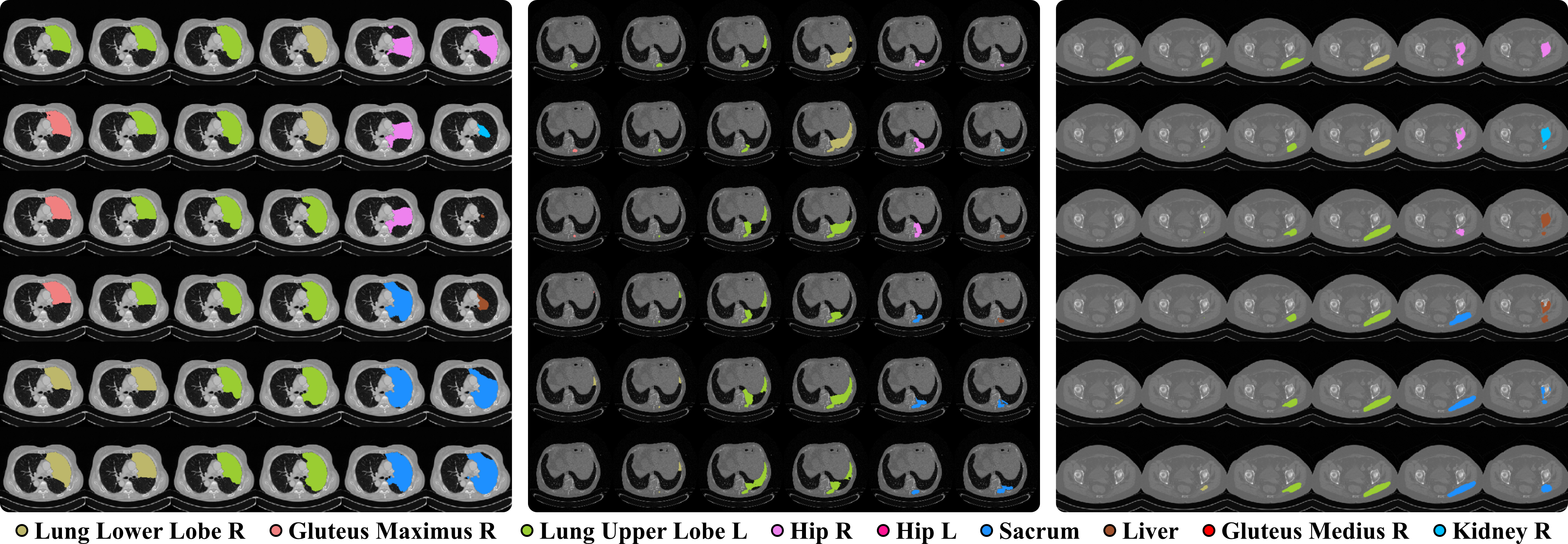}
    \caption{Grid of predictions generated by applying a fixed input image to a 6×6 grid of task vectors sampled uniformly from the spherical task space. Each output is colored according to the closest training class, revealing how different regions of the task space influence the segmentation behavior.}
    \label{fig:2dtaskvector}
\end{figure}

\subsection{Task Vector Interpolation}

To illustrate how the task vector influences segmentation, we interpolate between task vectors while keeping the input image and positional prompt fixed (Figure \ref{fig:taskvector_interpolation}). Interpolating from $liver \rightarrow kidney$, we observe the model initially producing no segmentation in non-liver regions. As the task vector moves toward the kidney representation, the model gradually segments kidney-like structures aligned to the spatial context. Further interpolation from $kidney \rightarrow \textit{lung lower lobe}$ shows intermediate states where the entire lung region is briefly highlighted before refining specifically to the lower lobe. To validate these dynamics, we sample task vectors uniformly from the unit sphere and visualize predictions on a fixed input image (Figure \ref{fig:2dtaskvector}). Each sampled vector is colored according to its nearest training class mean in task space, demonstrating \emph{distinct segmentation behaviors across different regions}. These transitions confirm that task vectors continuously guide predictions, encoding high-level shape and location priors that modulate segmentation dynamics.

\subsection{Limitations}
\label{sec:limitations}
While \mymethod{} achieves strong performance on seen tasks, its generalization to unseen targets remains below that of much larger specialized medical foundation models, partly due to the limited training data, just 4,500 slices across 15 tasks, compared to 11 million images used for training SAM. Training is also compute-intensive, requiring 93 hours in total (on an RTX 4090 GPU and AMD Ryzen 5700G CPU). A significant portion of this time stems from genetic optimization, with each round taking approximately 195 minutes. Currently, genetic and gradient-based optimization are performed in separate passes. In future work, unifying these passes could reduce training time and improve generalization, particularly when scaling to larger datasets.

\section{Conclusion}

This study demonstrates that foundation-model-like generalization can be achieved within the highly efficient framework of Neural Cellular Automata, paving the way for adaptive yet radically compact medical AI systems. By integrating gradient-based training with a novel \emph{genetic optimization strategy to evolve latent task representations}, we show that multi-task medical segmentation is feasible using \emph{only 211K parameters}, orders of magnitude fewer than traditional foundation models. Evaluated on 15 trained and 6 unseen segmentation tasks, \mymethod{} achieves an average Dice of 84.5\% on seen and 60.4\% on unseen targets, demonstrating \emph{meaningful task adaptation without retraining model weights}. These results highlight a new class of compact, prompt-driven segmentation models suitable for deployment in real-world, low-resource healthcare environments. By decoupling segmentation capabilities from parameter count and computational complexity, \mymethod{} lays the groundwork for accessible, reusable, and extensible medical AI. Rather than pursuing incremental performance with ever-larger models, we propose a shift toward compact, task-adaptive systems. This points toward a future where medical AI is defined not by infrastructure or cost, but by efficiency, flexibility, and universal accessibility.

\bibliographystyle{plainnat} 
\bibliography{references}

\appendix

\input{appendix}

\end{document}

%% file: appendix.tex
\section{Implementation Details}

This section outlines key aspects of the genetic optimization (GO) setup and training procedures used in \mymethod{}. All models are implemented in \texttt{PyTorch}, and genetic optimization is performed using a modified version of the \texttt{pygad} library, extended to support multiprocessing.

\subsection{Genetic Optimization During Training}

Genetic optimization is employed only during the \emph{first stage} of training, starting from epoch 20 and continuing through epoch 200.

\paragraph{GA Schedule and Parameters}
\begin{itemize}
    \item Task vector dimension: $3$
    \item Population size: $20$
    \item Initialization: Uniformly sampled from $[-1, 1]^3$, projected onto the unit sphere
    \item Gene bounds: $[-1, 1]$ (enforced via normalization)
    \item Update interval: Every 8 epochs, linearly increased to every 16
    \item Generations per GA step: $1$ (evaluate, generate offspring, evaluate, select best)
\end{itemize}

At each genetic step, 19 offspring are generated while the top candidate from the previous generation is retained. Fitness values are not cached between epochs, as the model evolves and caching would degrade performance.

\paragraph{Selection, Crossover, and Mutation}
\begin{itemize}
    \item Selection: Tournament selection with $k=3$, repeated $n=10$ times
    \item Crossover: Uniform crossover—each gene in the child is selected from one of two parents at random
    \item Mutation: Each gene is mutated with probability 0.1; mutation adds a random value from $[-1, 1]$
    \item Normalization: All offspring are projected back onto the unit sphere after mutation
\end{itemize}

\paragraph{Fitness Function}
During training, the inverse of the DiceBCE loss is used as the fitness value. DiceBCE is also used to train the NCA weights via gradient descent.

\subsection{Global Task Vector Optimization}

We also optimize \emph{global task vectors} for each class in two separate settings:

\paragraph{Training Tasks}
\begin{itemize}
    \item 20 epochs of GO
    \item Population size: $20$
    \item Initialization: Class mean + noise in $[-0.1, 0.1]$, clipped and normalized
    \item Fitness: Discrete Dice score
\end{itemize}

\begin{tcolorbox}[float=!t, title=Model Architecture: \texttt{\mymethod{}}, label={lst:model_config}]
\begin{lstlisting}
Foundational-NCA(
  (backbone_lowres): Backbone_Lowres(
    (p0): Conv2d(48, 48, kernel_size=(3, 3), stride=(1, 1), 
                 padding=(1, 1), padding_mode=reflect)
    (fc0): Linear(in_features=96, out_features=192, bias=True)
    (fc1): Linear(in_features=192, out_features=48, bias=False)
    (conditioning_after_p0): Sequential(
      (0): Conv2d(3, 192, kernel_size=(1, 1), stride=(1, 1))
      (1): SiLU()
      (2): Conv2d(192, 96, kernel_size=(1, 1), stride=(1, 1))
    )
    (conditioning_after_fc0): Sequential(
      (0): Conv2d(3, 192, kernel_size=(1, 1), stride=(1, 1))
      (1): SiLU()
      (2): Conv2d(192, 192, kernel_size=(1, 1), stride=(1, 1))
    )
  )
  (backbone_highres): Backbone_Highres(
    (p0): Conv2d(48, 48, kernel_size=(3, 3), stride=(1, 1), 
                 padding=(1, 1), padding_mode=reflect)
    (fc0): Linear(in_features=96, out_features=192, bias=True)
    (fc1): Linear(in_features=192, out_features=48, bias=False)
    (conditioning_after_p0): Sequential(
      (0): Conv2d(3, 192, kernel_size=(1, 1), stride=(1, 1))
      (1): SiLU()
      (2): Conv2d(192, 96, kernel_size=(1, 1), stride=(1, 1))
    )
    (conditioning_after_fc0): Sequential(
      (0): Conv2d(3, 192, kernel_size=(1, 1), stride=(1, 1))
      (1): SiLU()
      (2): Conv2d(192, 192, kernel_size=(1, 1), stride=(1, 1))
    )
  )
)
\end{lstlisting}
\captionof{figure}{Detailed PyTorch model configuration used for \mymethod{}.}
\end{tcolorbox}

\paragraph{Find Task Vector of Unknown Task}
\begin{itemize}
    \item 30 epochs of GO
    \item Population size: $40$
    \item Initialization: same as for training tasks
    \item Fitness: Discrete Dice score
\end{itemize}

\subsection{Model Training Setup}

We train \mymethod{} using the Adam optimizer with a learning rate of $1.6 \times 10^{-4}$ and $\beta$ parameters set to $(0.5, 0.5)$. An exponential learning rate scheduler with a decay rate of $\gamma = 0.9999$ is used throughout training.

Training is conducted in two phases: an initial phase of 200 epochs with alternating genetic optimization, followed by a second phase of 300 epochs using fixed task vectors. We use a batch size of 30 and a cell fire rate of $0.2$, meaning 80\% of cells are updated at each NCA step.

The architecture is implemented in PyTorch, and we detail the exact model structure in Figure~\ref{lst:model_config}. Initialization follows PyTorch defaults, except for two modifications: the input layer (\texttt{fc0}) is initialized using Xavier initialization, and the skip-connection layer (\texttt{fc1}) is initialized with a constant value of $0.001$.

\subsection{Hardware and System Setup}

All final experiments were conducted on a workstation with the following Hardware:

\begin{itemize}
    \item \textbf{CPU:} AMD Ryzen 7 5700G
    \item \textbf{RAM:} 60 GB
    \item \textbf{GPU:} NVIDIA RTX 4090 with 24 GB VRAM
    \item \textbf{Operating System:} Ubuntu 22.04.5 LTS
\end{itemize}

This setup was used for both model training and genetic optimization.

\begin{figure}[htbp]
    \centering
    \includegraphics[width=0.8\textwidth]{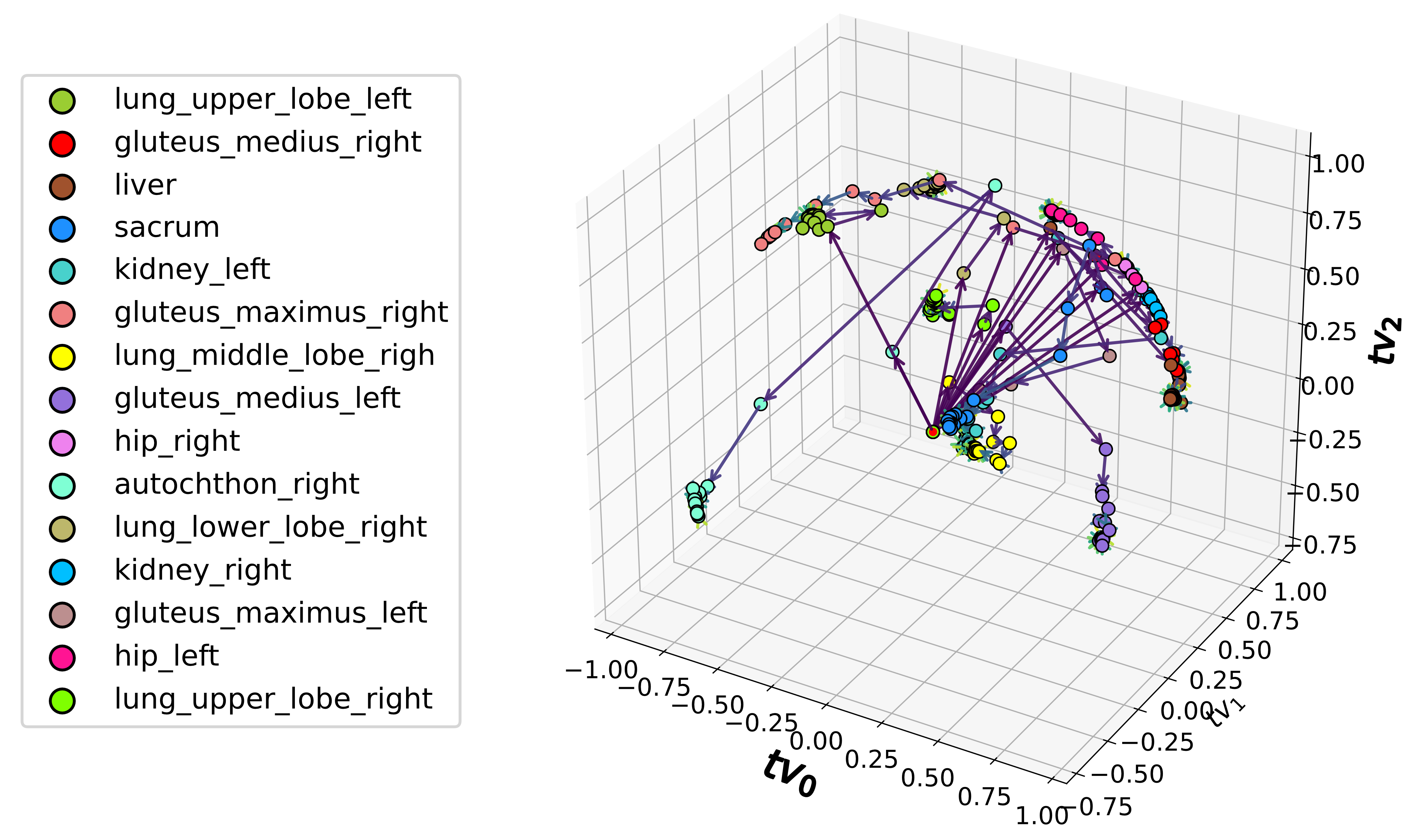}
    \caption{The mean task vector history shows early separation between tasks, which becomes increasingly refined with additional genetic optimization steps.}
    \label{fig:superTaskVectorHistory}
\end{figure}

\begin{figure}[htbp]
    \centering
    \includegraphics[width=0.8\textwidth]{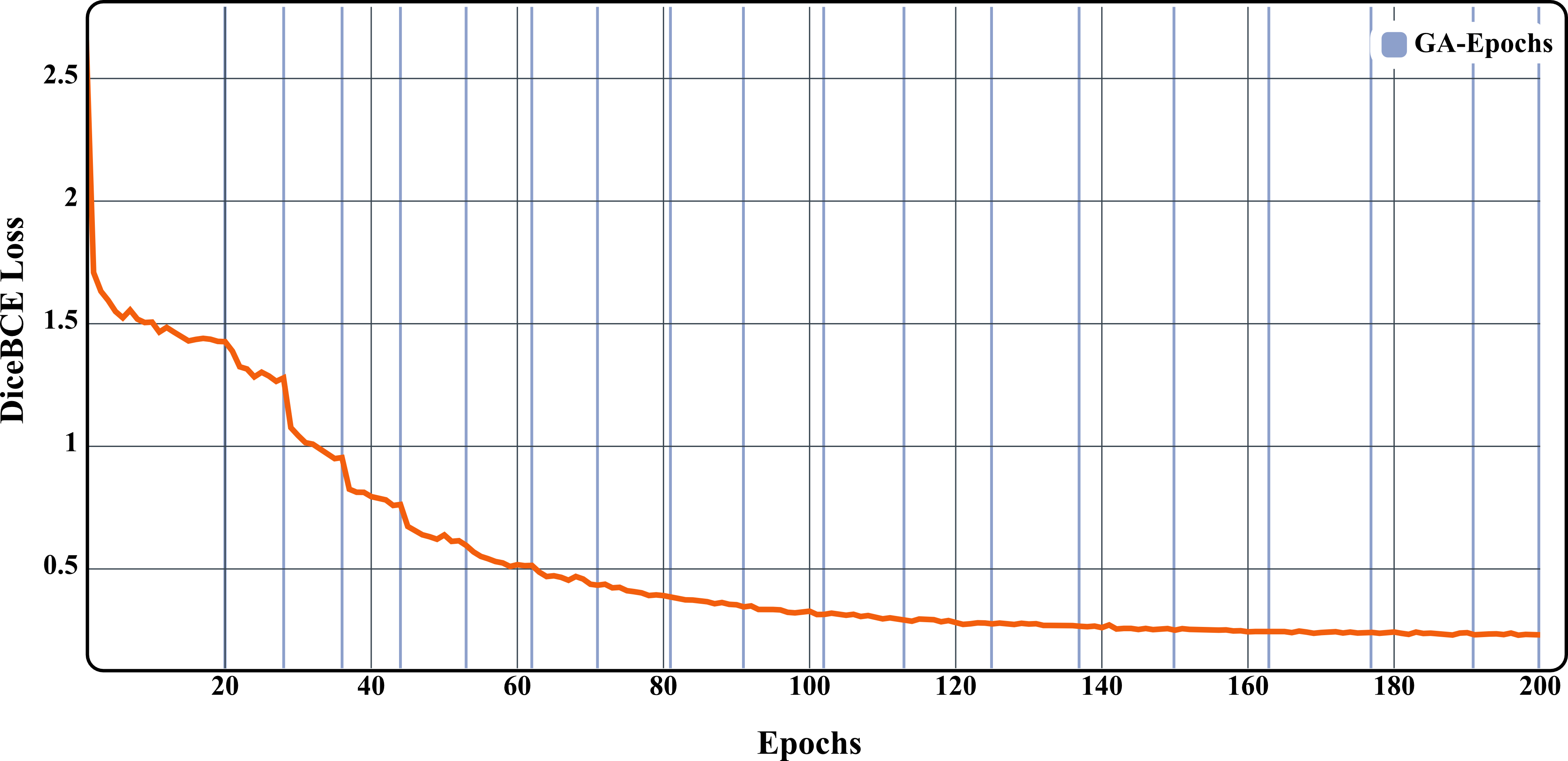}
    \caption{Loss trajectory of \mymethod{}. Clear improvements are observed following each genetic optimization step, especially during early training phases.}
    \label{fig:sup_losTrajectory}
\end{figure}

\begin{figure}[htbp]
    \centering
    \includegraphics[width=1.0\textwidth]{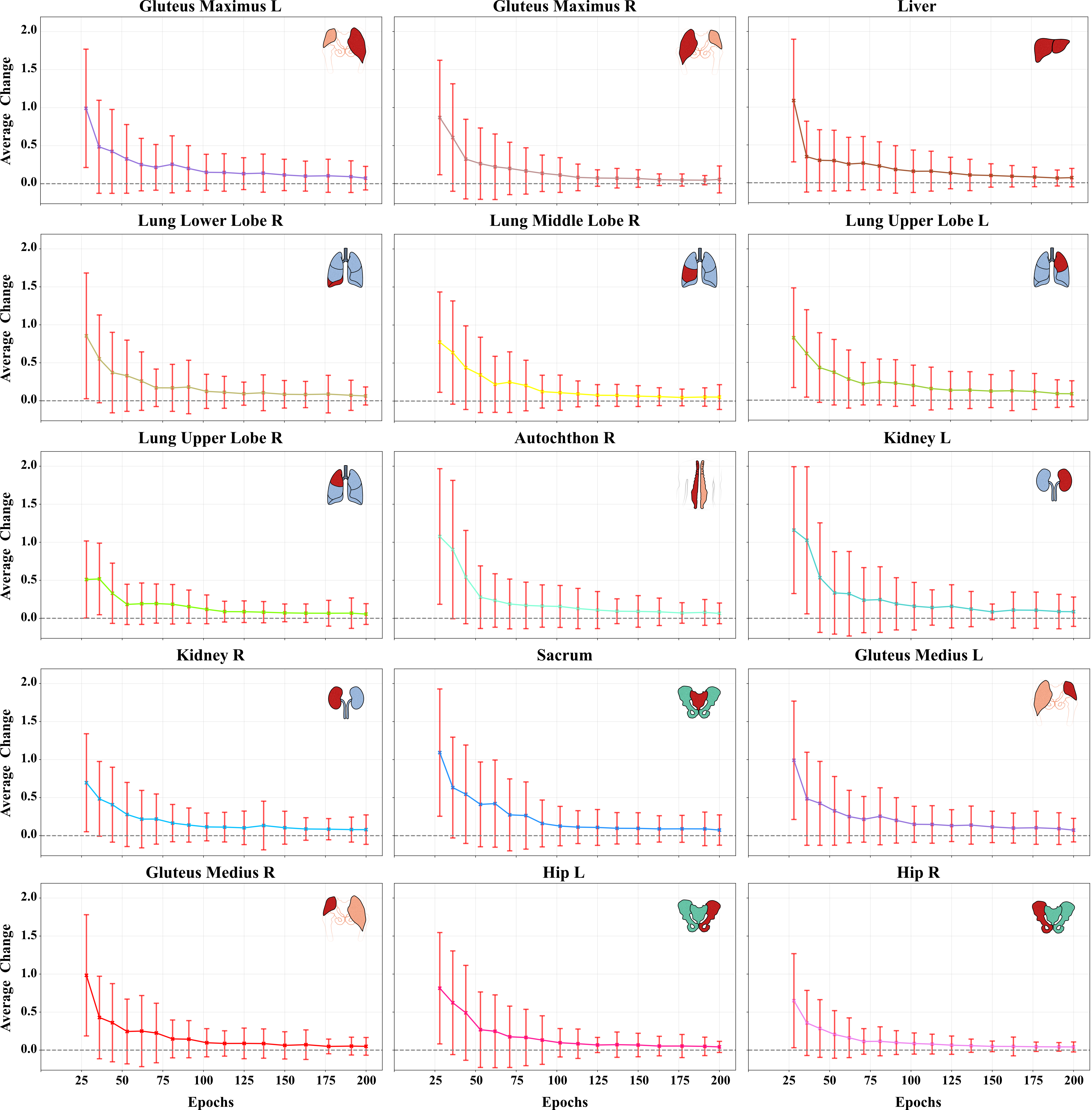}
    \caption{Genetic optimization trajectories for each label show consistent convergence toward an optimum, with some fluctuations remaining for more challenging segmentation targets.}
    \label{fig:sup_geneticChanges}
\end{figure}

\section{Training Trajectory}

To better understand how \mymethod{} evolves during training, we examine the optimization trajectory in more detail.

In Figure~\ref{fig:sup_losTrajectory}, we observe several phases of convergence throughout training. Notably, each round of genetic optimization leads to a distinct improvement in the loss, especially during the early epochs (e.g., around epoch 20) when the task space is still poorly structured. These early improvements highlight the importance of cold-start optimization in guiding the model toward more effective task representations.

Figure~\ref{fig:superTaskVectorHistory} further illustrates how the task space develops over time. Even in early stages, \mymethod{} begins to separate tasks into distinct regions on the unit sphere. Subsequent genetic updates primarily refine these groupings, though some tasks, such as Autochthon Right, undergo more significant adjustments. The figure shows median task vectors per class, projected onto the unit sphere for visualization.

Finally, Figure~\ref{fig:sup_geneticChanges} visualizes the per-class evolution of task vectors across training epochs. While most tasks exhibit stable convergence, others like Kidney Right show continued variation later in training, suggesting greater difficulty in anchoring those representations. In contrast, anatomies such as Hip Right demonstrate consistent early convergence. Importantly, this class-wise separation is used purely for analysis—\mymethod{} itself operates without access to class labels and must independently organize task representations based on supervision alone.

\section{Extended Ablation Study}

Due to space constraints in the main manuscript, we present here the full results of our ablation study evaluating the contributions of the positional and task prompts, as well as the impact of the training strategy. Table~\ref{tab:extended_ablation} provides detailed scores for both Task Set A and Task Set B across the complete set of conditions.

\paragraph{Single-Stage Setup: }

Beyond removing prompts individually, we investigate the effect of switching from our proposed two-stage training scheme to a simpler single-stage alternative. In the standard two-stage setup, we initially train with task prompts only, introducing positional prompts in the second stage to ensure both types of conditioning are fully utilized. In the single-stage baseline, both prompts are introduced at epoch 20.

Quantitatively, we observe that this joint conditioning from the start results in notably worse generalization on unseen tasks (Task Set B), with the average Dice score dropping from 0.604 to 0.572. Although performance on seen tasks (Task Set A) is comparable, the degraded generalization suggests that the task space becomes less structured when both prompts are introduced simultaneously.

\paragraph{Meaningful Task Separation: }

\begin{figure}[htbp]
    \centering
    \includegraphics[width=1.0\textwidth]{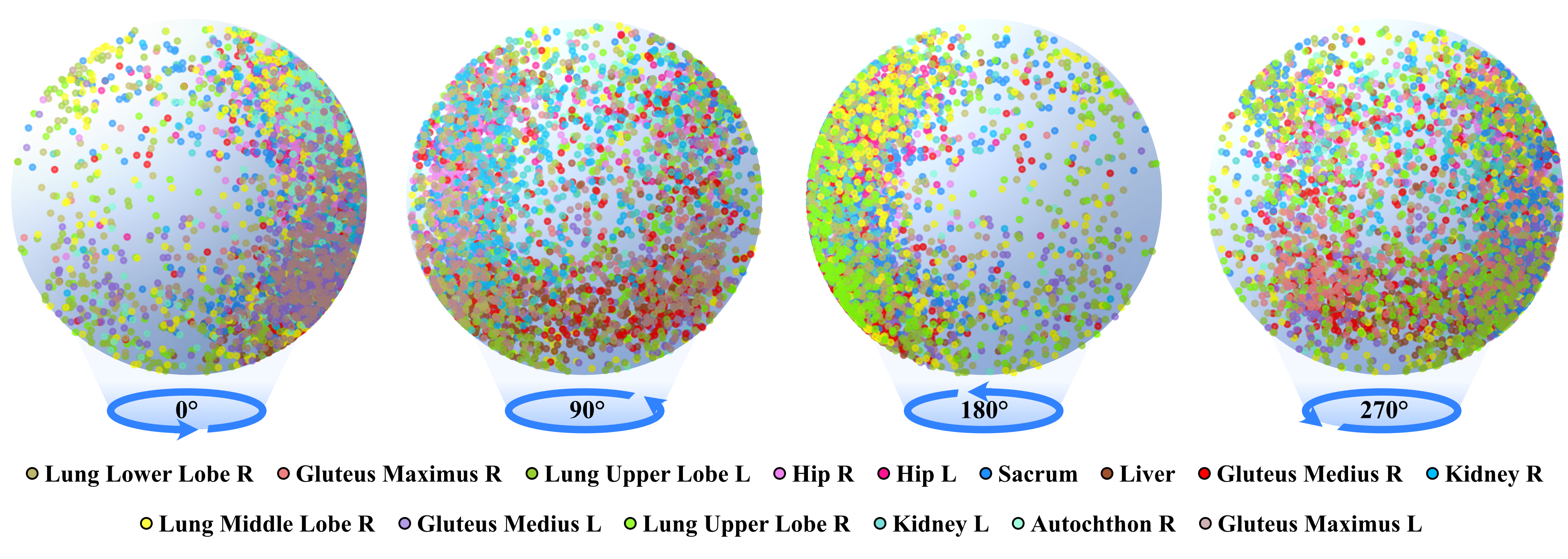}
    \caption{Task space visualization for the single-stage setup shows that introducing location and task information simultaneously leads to a disorganized and unstructured task representation.}
    \label{fig:supTaskSpaceAblation}
\end{figure}

This intuition is confirmed visually in Figure~\ref{fig:supTaskSpaceAblation}, where the task vectors learned under the single-stage setup show no meaningful separation or organization on the unit sphere. In contrast, the two-stage strategy encourages the formation of a structured, disentangled task space, which is critical for generalization.

\begin{figure}[htbp]
    \centering
    \includegraphics[width=1.0\textwidth]{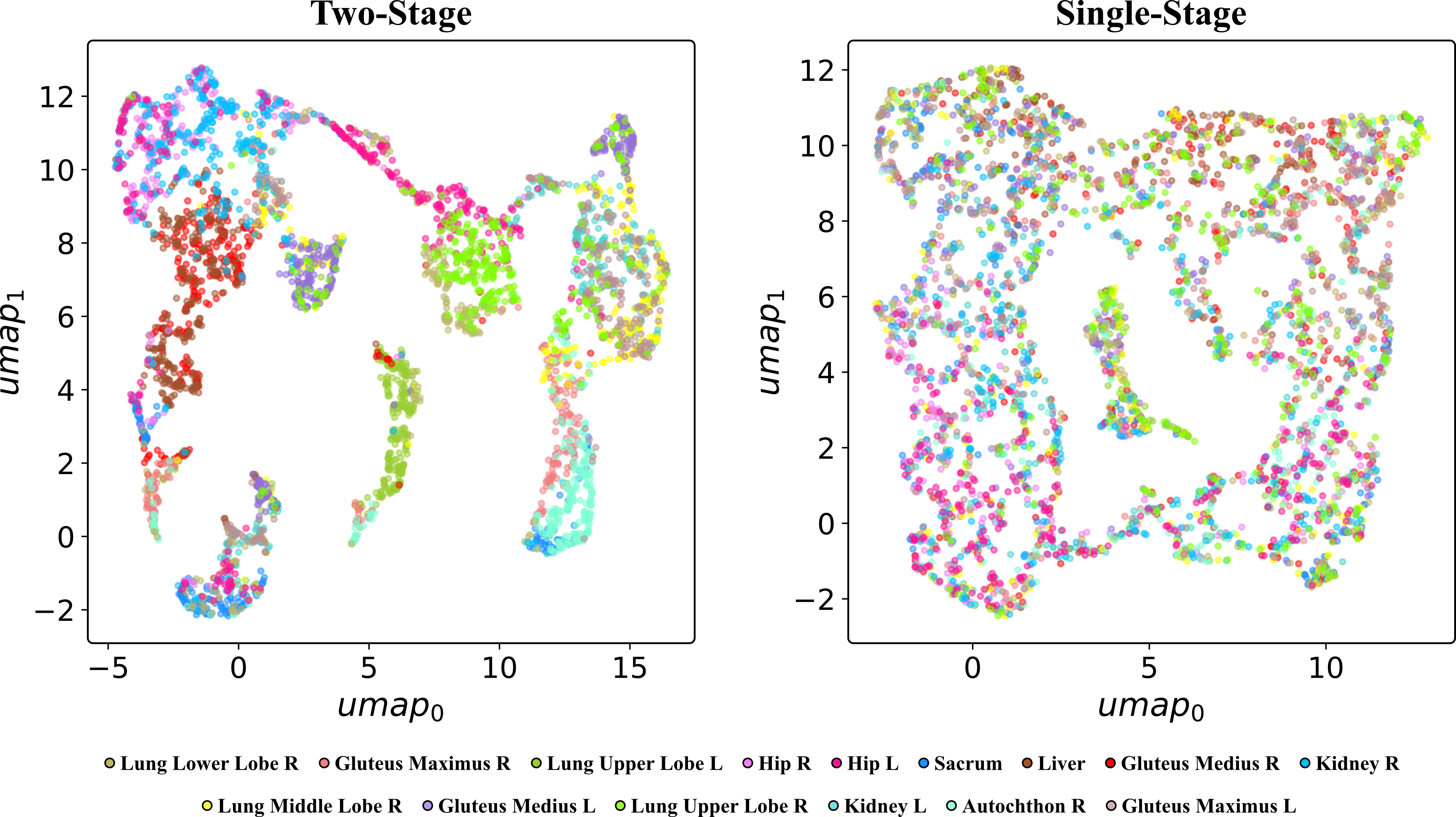}
    \caption{UMAP projection comparing single- and two-stage training. The two-stage setup leads to a more organized and interpretable task space, confirming earlier observations.}
    \label{fig:supUMAPProjection}
\end{figure}

Further evidence is provided in Figure~\ref{fig:supUMAPProjection}, which compares the UMAP projections of task vectors under both training strategies. The two-stage setup yields clearly clustered and well-separated task representations, while the single-stage approach results in overlapping and less distinguishable task regions, supporting our conclusion that careful prompt scheduling is crucial for robust, multi-task generalization.

\begin{table}[htbp]
\centering
\caption{Full Table of Ablation study of \mymethod{} on unseen tasks, evaluating the impact of removing the task vector and the positional prompt}
\begin{tabular}{lcccc}
\toprule
\textbf{Task Set A} & \textbf{\mymethod{}} & \textbf{No Location} & \textbf{No Task} & \textbf{Single Stage} \\
\toprule
Gluteus Max. L & 0.885 $\pm$ 0.077 & 0.867 $\pm$ 0.114 & 0.879 $\pm$ 0.085 & 0.879 $\pm$ 0.090 \\
Gluteus Max. R & 0.874 $\pm$ 0.073 & 0.807 $\pm$ 0.131 & 0.876 $\pm$ 0.089 & 0.877 $\pm$ 0.098 \\
Liver & 0.911 $\pm$ 0.105 & 0.874 $\pm$ 0.170 & 0.892 $\pm$ 0.116 & 0.904 $\pm$ 0.122 \\
Lung Lower Lobe R & 0.832 $\pm$ 0.162 & 0.763 $\pm$ 0.188 & 0.886 $\pm$ 0.139 & 0.813 $\pm$ 0.173\\
Lung Middle Lobe R & 0.808 $\pm$ 0.114 & 0.736 $\pm$ 0.143 & 0.837 $\pm$ 0.084 & 0.825 $\pm$ 0.106 \\
Lung Upper Lobe L & 0.870 $\pm$ 0.170 & 0.786 $\pm$ 0.205 & 0.887 $\pm$ 0.156 & 0.841 $\pm$ 0.186 \\
Lung Upper Lobe R & 0.800 $\pm$ 0.271 & 0.732 $\pm$ 0.266 & 0.854 $\pm$ 0.225 & 0.806 $\pm$ 0.267 \\
Autochthon R & 0.811 $\pm$ 0.060 & 0.681 $\pm$ 0.146 & 0.804 $\pm$ 0.067 & 0.814 $\pm$ 0.063 \\
Kidney L & 0.840 $\pm$ 0.186 & 0.706 $\pm$ 0.299 & 0.858 $\pm$ 0.168 & 0.854 $\pm$ 0.178 \\
Kidney R & 0.871 $\pm$ 0.153 & 0.651 $\pm$ 0.291 & 0.847 $\pm$ 0.146 & 0.875 $\pm$ 0.140 \\
Sacrum & 0.758 $\pm$ 0.104 & 0.691 $\pm$ 0.138 & 0.719 $\pm$ 0.105 & 0.695 $\pm$ 0.111 \\
Gluteus Medius L & 0.828 $\pm$ 0.124 & 0.819 $\pm$ 0.139 & 0.804 $\pm$ 0.118 & 0.788 $\pm$ 0.112 \\
Gluteus Medius R & 0.840 $\pm$ 0.047 & 0.780 $\pm$ 0.124 & 0.815 $\pm$ 0.102 & 0.821 $\pm$ 0.083 \\
Hip L & 0.863 $\pm$ 0.101 & 0.665 $\pm$ 0.136 & 0.876 $\pm$ 0.102 & 0.881 $\pm$ 0.101 \\
Hip R & 0.887 $\pm$ 0.104 & 0.735 $\pm$ 0.135 & 0.881 $\pm$ 0.103 & 0.895 $\pm$ 0.102 \\
\toprule
Average & 0.845 & 0.753 & 0.848 & 0.838\\
\midrule
\textbf{Task Set B} & \textbf{\mymethod{}} & \textbf{No Location} & \textbf{No Task} & \textbf{Single Stage} \\
\midrule
Autochthon L & 0.678 $\pm$ 0.159 & 0.002 $\pm$ 0.012 & 0.661 $\pm$ 0.148 & 0.624 $\pm$ 0.169 \\
Iliopsoas L & 0.568 $\pm$ 0.174 & 0.020 $\pm$ 0.041 & 0.593 $\pm$ 0.141 & 0.536 $\pm$ 0.172 \\
Lung Lower Lobe L & 0.730 $\pm$ 0.153 & 0.299 $\pm$ 0.168 & 0.746 $\pm$ 0.158 & 0.749 $\pm$ 0.158\\
Sternum & 0.661 $\pm$ 0.088 & 0.018 $\pm$ 0.038 & 0.645 $\pm$ 0.079 & 0.705 $\pm$ 0.065\\
Stomach & 0.492 $\pm$ 0.184 & 0.019 $\pm$ 0.046 & 0.392 $\pm$ 0.152 & 0.394 $\pm$ 0.177 \\
Urinary Bladder & 0.497 $\pm$ 0.179 & 0.036 $\pm$ 0.058 & 0.271 $\pm$ 0.139 & 0.423 $\pm$ 0.160\\
\midrule
Average & 0.604 & 0.066 & 0.551 & 0.572\\

\bottomrule
\end{tabular}
\label{tab:extended_ablation}
\end{table}

\section{Additional Baselines of SAM-H and SAM2.1-L}

To complement our main comparisons, we additionally evaluate two high-parameter SAM variants: SAM-H (641M parameters) and SAM2.1-L (224M parameters). As shown in Table~\ref{tab:add_baselines}, both models perform similarly to their smaller counterparts, achieving average Dice scores of 0.289 and 0.305 on Task Set A, and 0.173 and 0.175 on Task Set B, respectively. These results suggest that simply increasing model size does not yield significant improvements in medical segmentation tasks, highlighting the efficiency of compact, task-adaptive alternatives such as \mymethod{}.

\begin{table}[htbp]
\centering
\caption{Additional Baseline results of the high parameter variants of SAM and SAM2.}
\begin{tabular}{lcc}
\toprule
\textbf{Task Set A} & \textbf{SAM-H} & \textbf{SAM2.1-L}  \\
\emph{Parameters} & \emph{641.1M} & \emph{224.4M}\\
\toprule
Gluteus Max. L &  0.076 $\pm$ 0.020  & 0.074 $\pm$ 0.021 \\
Gluteus Max. R &  0.074 $\pm$ 0.019  & 0.077 $\pm$ 0.022 \\
Liver &  0.367 $\pm$ 0.089  & 0.443 $\pm$ 0.124 \\
Lung Lower Lobe R &  0.583 $\pm$ 0.176  & 0.532 $\pm$ 0.171 \\
Lung Middle Lobe R &  0.349 $\pm$ 0.122  & 0.305 $\pm$ 0.106 \\
Lung Upper Lobe L &  0.497 $\pm$ 0.223  & 0.398 $\pm$ 0.197 \\
Lung Upper Lobe R &  0.535 $\pm$ 0.215  & 0.495 $\pm$ 0.204 \\
Autochthon R &  0.031 $\pm$ 0.010  & 0.036 $\pm$ 0.011 \\
Kidney L &  0.389 $\pm$ 0.278  & 0.553 $\pm$ 0.301 \\
Kidney R &  0.401 $\pm$ 0.268 & 0.562 $\pm$ 0.295 \\
Sacrum &  0.187 $\pm$ 0.109  & 0.199 $\pm$ 0.134 \\
Gluteus Medius L & 0.052 $\pm$ 0.014 & 0.062 $\pm$ 0.084 \\
Gluteus Medius R & 0.048 $\pm$ 0.011 & 0.051 $\pm$ 0.034 \\
Hip L & 0.349 $\pm$ 0.180 & 0.402 $\pm$ 0.192 \\
Hip R & 0.397 $\pm$ 0.189 & 0.391 $\pm$ 0.197 \\
\toprule
Average & 0.289 & 0.305\\
\midrule
\textbf{Task Set B} & \textbf{SAM-H} & \textbf{SAM2.1-L}  \\
\emph{Parameters} & \emph{641.1M} & \emph{224.4M}\\
\midrule
Autochthon L & 0.031 $\pm$ 0.010 & 0.036 $\pm$ 0.012 \\
Iliopsoas L & 0.027 $\pm$ 0.017 & 0.028 $\pm$ 0.020 \\
Lung Lower Lobe L & 0.531 $\pm$ 0.204 & 0.461 $\pm$ 0.206 \\
Sternum & 0.178 $\pm$ 0.207 & 0.135 $\pm$ 0.183 \\
Stomach & 0.126 $\pm$ 0.067 & 0.193 $\pm$ 0.135 \\
Urinary Bladder & 0.141 $\pm$ 0.114 & 0.198 $\pm$ 0.150 \\
\midrule
Average & 0.173 & 0.175 \\

\bottomrule
\end{tabular}
\label{tab:add_baselines}
\end{table}

\section{Additional Qualitative Results}

\begin{figure}[htbp]
    \centering
    \includegraphics[width=1.0\textwidth]{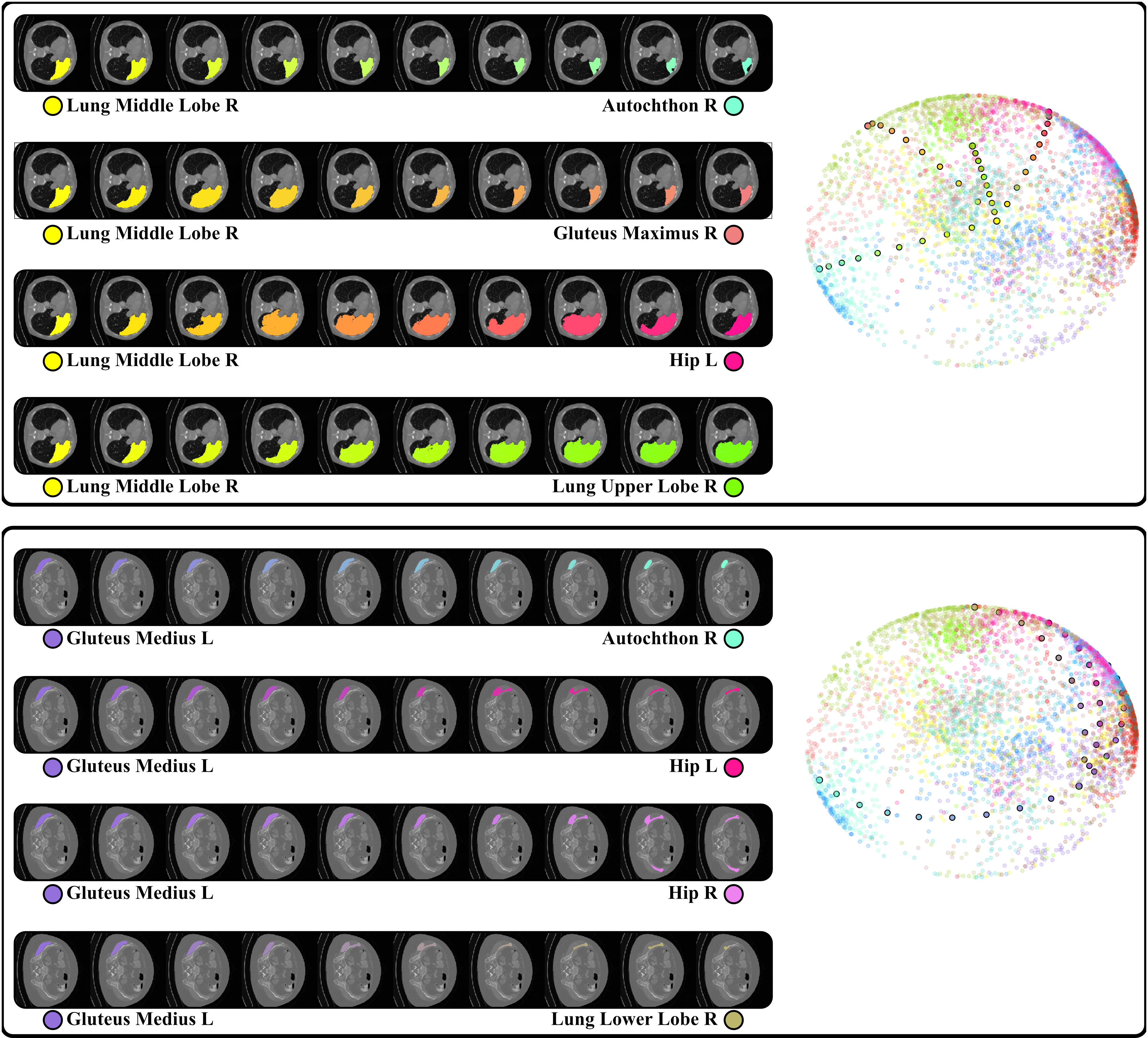}
    \caption{Additional task vector interpolations further confirm that the model adapts the segmentation to fit the task prompt as well as possible within the spatial constraints given by the positional prompt.}
    \label{fig:superAdditionalInterpolations}
\end{figure}

\paragraph{Task Vector Interpolations:} To further explore the influence of the task prompt on the model’s predictions, we visualize interpolations from a single starting task vector to four distinct target vectors in Figure \ref{fig:superAdditionalInterpolations}. This illustrates how the segmentation output evolves as the task embedding changes. For example, starting from the vector for \emph{Lung Middle Lobe R}, we observe distinct and substantial changes when interpolating toward \emph{Autochthon R} and \emph{Lung Upper Lobe R}, highlighting the model’s sensitivity to semantic shifts in task space. A similar pattern emerges when starting from \emph{Gluteus Medius L}, in this case, interpolation toward \emph{Hip R} even causes the model to segment the anatomically opposite side, demonstrating how certain transitions in task space can induce asymmetric responses.

\paragraph{Sampling from the Unit Sphere:} We extend the task vector sampling experiment introduced in the main manuscript. In Figure \ref{fig:sup_MoreInterpolationsEvenlySampled}, we sample task vectors uniformly from the unit sphere and visualize their corresponding outputs on two distinct images in the color of the closest average training anatomy. This larger-scale evaluation reveals a wide range of segmentation behaviors depending on the location in the task space. Importantly, we find that the positional prompt also plays a key role: in Sample 1 (left), where the positional prompt anchors the expected location of the anatomy, the model produces more consistent predictions across task vectors. In contrast, Sample 2 (right), exhibits more pronounced variability depending on the task vector, emphasizing the interplay between task and spatial conditioning.

\paragraph{Additional Qualitative Results:} Finally, we provide an extended set of qualitative results in Figure \ref{fig:sup_MoreInterpolationsEvenlySampled}, covering all 15 training tasks and 6 unseen tasks. On the seen tasks, \mymethod{} demonstrates robust segmentation performance, often capturing fine anatomical details and accommodating large structural variation, for instance, the liver’s varying morphology. For unseen tasks, performance varies depending on how well the new class is represented in the learned task space. While certain classes like \emph{Hip R} are segmented with high fidelity, others such as \emph{Lung Lower Lobe L} show inconsistencies, especially in more complex regions. These findings underscore the model’s capacity for generalization, while also highlighting the challenges posed by underrepresented or highly variable anatomies. This limitation is likely exacerbated by the relatively small training set of 4,500 annotated slices across 15 tasks. In future work, increasing the number and diversity of training samples, particularly for anatomically complex or rare structures, could substantially enrich the learned task space, leading to more robust adaptation and improved segmentation accuracy across a broader range of unseen targets.

\begin{sidewaysfigure}[htbp]
    \centering
    \includegraphics[width=1.0\textwidth]{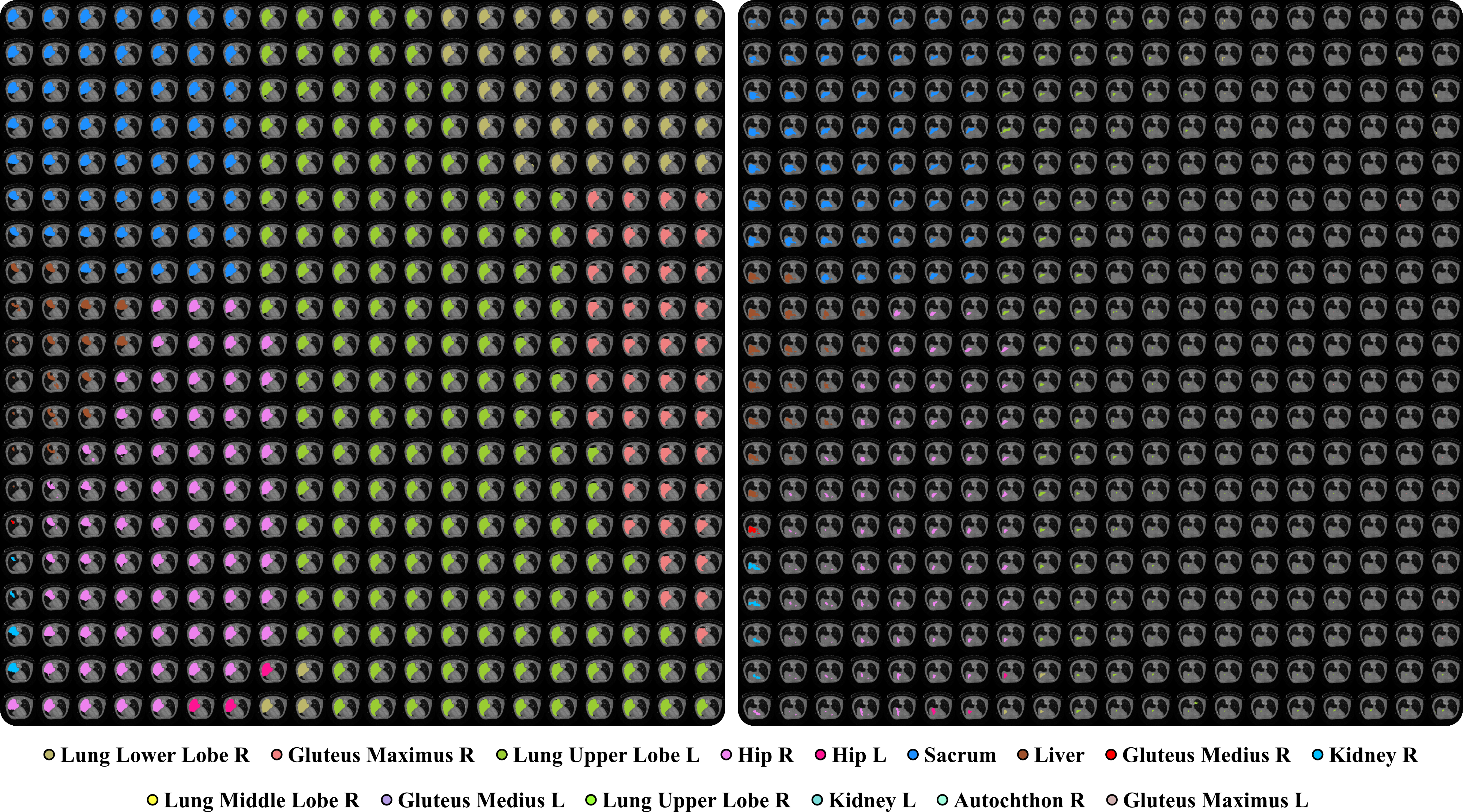}
    \caption{Evenly sampled task prompts from the unit sphere, applied to two input images with corresponding positional prompts. Each output is labeled according to the closest mean class in the task space.}
    \label{fig:sup_MoreInterpolationsEvenlySampled}
\end{sidewaysfigure}

\begin{sidewaysfigure}[htbp]
    \centering
    \includegraphics[width=1.0\textwidth]{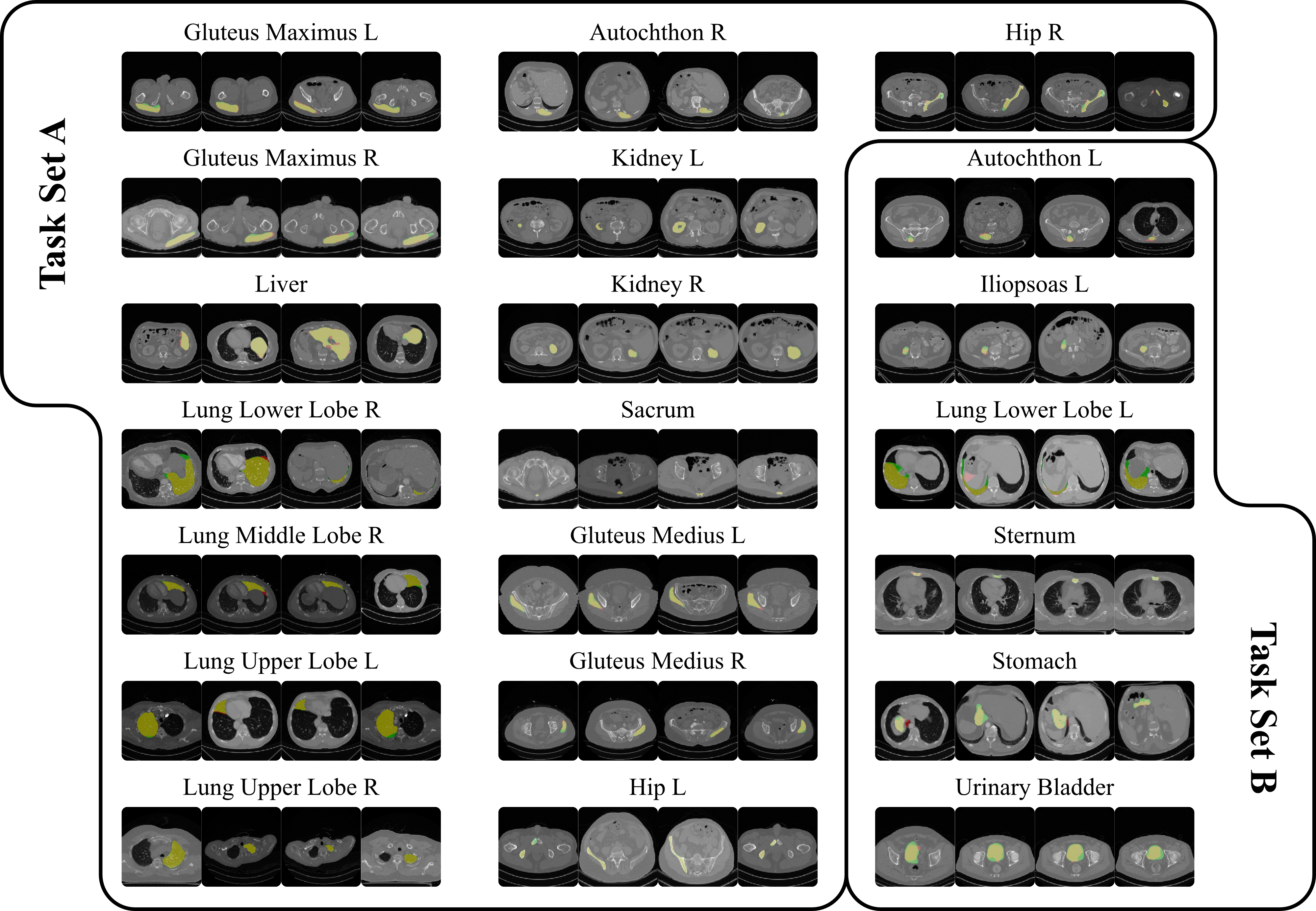}
    \caption{Qualitative results for all 15 training tasks (Task Set A) and 6 unseen tasks (Task Set B). Green: ground truth, Red: prediction, Yellow: overlap.}
    \label{fig:sup_MoreInterpolationsEvenlySampled}
\end{sidewaysfigure}